\documentclass[a4paper]{article}
\usepackage[T1]{fontenc}
\usepackage{graphicx}
\usepackage{booktabs}

\usepackage{tikz}
\usetikzlibrary{arrows.meta,positioning,fit,calc,shapes.symbols,shapes.geometric}

\usepackage{makecell}
\newcommand{\rot}[1]{%
  \makecell[b]{\rotatebox[origin=lb]{90}{\strut #1}}%
}

\makeatletter
\@ifundefined{doi}
  {\newcommand{\doi}[1]{\href{https://doi.org/#1}{\nolinkurl{#1}}}}
  {\renewcommand{\doi}[1]{\href{https://doi.org/#1}{\nolinkurl{#1}}}}
\makeatother

\usepackage[hidelinks]{hyperref}
\usepackage{amsfonts}

\usepackage{authblk}

\begin{document}

\title{Data-Local Autonomous LLM-Guided Neural Architecture Search for Multiclass Multimodal Time-Series Classification}

\author[1]{Emil Hardarson\thanks{Corresponding author: emilh@ru.is}}
\author[1]{Luka Biedebach}
\author[1]{Ómar Bessi Ómarsson}
\author[1]{Teitur Hrólfsson}
\author[1]{Anna Sigridur Islind}
\author[1,2]{María Óskarsdóttir}

\affil[1]{Reykjavik University, Reykjavik, Iceland}
\affil[1]{University of Southampton, Southampton, United Kingdom}

\date{}

\maketitle 

\begin{abstract}
Applying machine learning to sensitive time-series data is often bottlenecked by the iteration loop: Performance depends strongly on preprocessing and architecture, yet training often has to run on-premise under strict data-local constraints. This is a common problem in healthcare and other privacy-constrained domains (e.g., a hospital developing deep learning models on patient EEG). This bottleneck is particularly challenging in multimodal fusion, where sensor modalities must be individually preprocessed and then combined. LLM-guided neural architecture search (NAS) can automate this exploration, but most existing workflows assume cloud execution or access to data-derived artifacts that cannot be exposed.

We present a novel data-local, LLM-guided search framework that handles candidate pipelines remotely while executing all training and evaluation locally under a fixed protocol. The controller observes only trial-level summaries, such as pipeline descriptors, metrics, learning-curve statistics, and failure logs, without ever accessing raw samples or intermediate feature representations. Our framework targets multiclass, multimodal learning via one-vs-rest binary experts per class and modality, a lightweight fusion MLP, and joint search over expert architectures and modality-specific preprocessing.

We evaluate our method on two regimes: UEA30 (public multivariate time-series classification dataset) and SleepEDFx sleep staging (heterogeneous clinical modalities such as EEG, EOG, and EMG). 
The results show that the modular baseline model is strong, and the LLM-guided NAS further improves it. Notably, our method finds models that perform within published ranges across most benchmark datasets. Across both settings, our method reduces manual intervention by enabling unattended architecture search while keeping sensitive data on-premise.


\end{abstract}

\section{Introduction}

\paragraph{Motivation.}
Designing preprocessing pipelines and model architectures for classifying multiclass, multimodal time-series data is a challenge for machine learning engineers. Multimodal time series data refers to multiple synchronized streams of temporal data of different modalities \cite{baltruvsaitis2018multimodal}. These temporal modalities may include, for example, sensor measurements, audio, video, or physiological data \cite{lahat2015multimodal}. With the increasing presence of smartphones, digitalized manufacturing, wearable devices, smart home sensors, and medical devices, an abundance of multimodal data is frequently produced and is becoming increasingly complex \cite{Ghosal2021Multi}. Potential uses of this data include healthcare \cite{El2020Multimodal}, activity recognition \cite{Li2016Multi}, and the Internet of Things \cite{Inan2025DeepMetaIoT}. Physiological sleep monitoring is a prime example of multimodal time-series classification, as it involves synchronized signals from different modalities \cite{chambon2018deep}, which is why we use it as an application domain for our proposed system.

With increasing numbers of modalities, experimentation becomes more complex. They pose a particularly challenging design space because (i) modalities are heterogeneous and provide complementary evidence, (ii) class structure induces non-uniform confusion patterns, and (iii) architectural choices interact with modality handling and fusion strategy \cite{baltruvsaitis2018multimodal}. Consequently, the standard workflow of manually iterating on architectures and training protocols can be both slow and brittle. AutoML tools, which automate feature engineering, model selection, hyperparameter tuning, and evaluation, promise a way to overcome this challenge \cite{karmaker2021automl}. Particularly, Neural Architecture Search (NAS), introduced in 2016 by Zoph and Le, specifically searches the architecture space of artificial neural networks \cite{zoph2016neural}.

\paragraph{Problem Statement.}
 Nowadays, NAS and related automated machine learning techniques have become standard tools for reducing the manual effort of model design. However, in many applied settings, the dominant bottleneck is not the absence of optimization primitives, but the practical burden of \emph{running} long-horizon experimentation: configuring and restarting runs, diagnosing failures, tracking intermediate artifacts, and deciding what to try next under limited compute resources. Furthermore, many domains that produce multimodal data, such as healthcare, are prohibited from applying these tools, since sensitive, personal data cannot be moved to an external tool or server for optimization and training. In such settings, progress is often limited by \emph{human-hours per unit of experimental progress}, even when substantial compute time is acceptable.

\paragraph{Proposed Solution.}
To address these challenges, we propose an autonomous, data-local experimentation framework that formalizes NAS-driven experimentation as a \emph{research cycle}: the system repeatedly proposes candidate expert architectures, instantiates training runs, evaluates outcomes, and records results as persistent artifacts. 
The framework is designed to reduce the number of human hours spent on orchestration and iteration while allowing long-running, compute-intensive search processes to run unattended. To make the multiclass multimodal structure explicit, we adopt a modular decomposition in which we train one binary expert per class and per modality, and then learn a lightweight fusion model over the experts' penultimate representations. 

Another novelty of our proposed solution is a large-language-model (LLM)-guided optimization approach. We employ an LLM as a controller to guide the neural architecture search for the individual expert models, proposing candidate architectures based on observed validation performance and training dynamics from prior evaluations. All model training and evaluation are performed locally under a fixed protocol, with the LLM receiving only aggregated metrics and artifacts, enabling iterative architecture optimization while preserving data locality and privacy.

\paragraph{Contributions.} 
We present and evaluate the proposed framework and provide a proof-of-concept in two settings. First, by comparing a simple baseline NAS with existing classifiers in UEA30, a benchmarking dataset of multimodal classification tasks, and sleepEDFx, a sleep classification benchmarking data set. Second, by showing the enhanced performance of the LLM-guided NAS. 
Existing solutions have shown to be effective for one-dimensional data, but multimodal time series data has, to the best of our knowledge, not been approached with AutoML tools yet. 
In summary, our main contributions are the following:
\begin{itemize}
    \item We introduce an autonomous, data-local experimentation framework for NAS that reduces human-hours by formalizing experimentation as a research cycle with persistent, auditable artifacts.
    \item We operationalize multiclass multimodal learning via a modular expert decomposition: one binary classifier per class and per modality, with a trainable fusion MLP over penultimate representations; NAS is applied only to the experts.
    \item We provide empirical evidence on 22 multiclass multivariate time-series tasks and demonstrate the effectiveness on our approach on two of them.
\end{itemize}

The rest of this paper is organized as follows. In the next section, we discuss research that is related to our work. In Section \ref{sec:method}, we present the two parts of our proposed methodology: the fusion network of binary experts and the LLM-guided NAS. Section \ref{sec:results} presents the results of our experiments, and in Section \ref{sec:discussion} we discuss their implications, e.g. regarding data privacy. Finally, we summarize our work and contributions in Section \ref{sec:conclusion}.

\section{Related Work}

Various AutoML \cite{zoller2021benchmark} and NAS \cite{ren2021comprehensive} tools have been proposed in existing literature. Our contribution to this existing body of work is by combining several research threads: multimodal time-series classification and fusion, multiclass decomposition and expert-style modeling, NAS for temporal and multimodal architectures, practical AutoML and long-horizon experimentation, and the emerging use of LLMs as assistants in machine learning workflows. The unifying motivation is to support systematic model exploration for multiclass multimodal time series under realistic operational constraints, with an emphasis on reducing human effort in iterative experimentation.

\paragraph{AutoML for Ensemble Learning.} 
A central design dimension in multimodal learning is the fusion strategy: early fusion combines modalities at the input or low-level feature stage, intermediate fusion combines learned representations (often through concatenation, attention, or gating), and late fusion combines decisions (e.g., logits or probabilities) produced by modality-specific models \cite{tavakoli2025multi}. We propose an architecture that creates separate models for each modality, which is in itself not a novel approach, as it has been applied, for example, by Bastos et al. \cite{Bastos2025Multimodal}, but we add to existing literature by developing a system to automatically optimize these separate models. Erickson et al. introduced AutoGluon, an AutoML tool that automatically optimizes the building of multiple ensemble models \cite{erickson2020autogluon}. McKinnon and Atkinson developed a search space for multivariate time-series NAS \cite{MacKinnon2023Designing}, but no existing tool addresses the specific needs of multimodal time-series data. 

\paragraph{LLM-guided AutoML.} 
A recent thread in AutoML explores using LLMs to assist or guide NAS, motivated by the observation that many architecture changes are naturally expressed, for example, as structured edits to code or high-level design constraints. The natural language used in these models is a fundamental difference to traditional AutoML tools, as Miguel-Morante et al. showed with their AutoML tool, emphasizing the human-centric approach of providing ML systems in a more intuitive and accessible way and strengthening the role of human-computer interaction \cite{miguel2025integrating}. Existing LLM-to-NAS methods vary in how they control representation and feedback. Trirat, Jeong, and Hwang, for example, created an agent-based LLM-guided AutoML tool that takes users’ task descriptions and facilitates collaboration among specialized LLM agents to deliver deployment-ready models \cite{trirat2024automl}.

\paragraph{Data-Local AutoML.} 
Our work adds to existing research and existing AutoML software, and, furthermore, from an data-privacy perspective. Since our architecture allows data to be stored locally, improving the model iteratively solely by receiving textual summary information about the data and reviewing training statistics ensures full data privacy and protection. Other AutoML tools, such as H2O \cite{ledell2020h2o} and AutoSklearn \cite{feurer2022auto}, support a fully local implementation, which also ensures privacy. Our solution, however, preserves this local property while still leveraging the capabilities of remotely hosted LLMs, which would not be feasible in a fully local implementation.

\section{Method}\label{sec:method}

\subsection{Fusion network of modular binary experts}\label{sec:expert}
We consider the problem of multiclass classification of multimodal time-series observations. Let $m \in \{1,\dots,M\}$ be an index for modalities (e.g., sensor channels), and let $y \in \{1,\dots,N\}$ denote the class label with $c\in\{1,\dots, C\}$ classes. Each example consists of modality-specific time series $(x_1,\dots,x_M)$, where $x_m \in \mathbb{R}^{T_m \times d_m}$ may have modality-dependent length $T_m$ and dimensionality $d_m$. The goal is to learn a predictor $\hat{y} = F(x_1,\dots,x_M)$ that maps multimodal time series to a multiclass label.

\begin{figure}[htbp]
\centering
\resizebox{0.95\textwidth}{!}{%
\begin{tikzpicture}[
  font=\rmfamily,
  >=Latex,
  lab/.style={},
  expert/.style={draw, thick, rounded corners=2pt, minimum width=18mm, minimum height=7mm, align=center},
  rep/.style={draw, thick, rounded corners=2pt, minimum width=18mm, minimum height=6mm, align=center},
  vec2/.style={draw, thick, minimum width=2mm, minimum height=4mm, inner sep=0pt},
  fusion/.style={draw, thick, rounded corners=3pt, minimum width=12mm, minimum height=80mm, align=center},
  vec5/.style={draw, thick, minimum width=2mm, minimum height=10mm, inner sep=0pt}
]

\node[lab, align=center] at (0,1.2)   {Binary\\experts};
\node[lab, align=center] at (5.7,1.2) {Penultimate\\representations};
\node[lab] at (8.0,1.2) {Fusion};


\node[lab, anchor=east] (sig1) at (-2.0,-1.1) {Signal 1};
\node                   (sigdots) at (-2.0,-4.9) {$\vdots$};
\node[lab, anchor=east] (sigM) at (-2.0,-6.75) {Signal M};

\node[expert] (f11) at (0,0)    {$f_{1,1}$};
\node[expert] (f12) at (0,-1.1) {$f_{1,2}$};
\node         (f1vd) at (0,-2.2) {$\vdots$};
\node[expert] (f1N) at (0,-3.3) {$f_{1,N}$};

\node (fvd) at (0,-4.9) {$\vdots$};

\node[expert] (fMNm1) at (0,-6.2) {$f_{M,N-1}$};
\node[expert] (fMN)   at (0,-7.3) {$f_{M,N}$};

\draw[thick, shorten <=1.0mm, shorten >=1.0mm] (sig1.east) -- (f11.west);
\draw[thick, shorten <=1.0mm, shorten >=1.0mm] (sig1.east) -- (f12.west);
\draw[thick, shorten <=1.0mm, shorten >=1.0mm] (sig1.east) -- (f1N.west);

\draw[thick, shorten <=1.0mm, shorten >=1.0mm] (sigM.east) -- (fMNm1.west);
\draw[thick, shorten <=1.0mm, shorten >=1.0mm] (sigM.east) -- (fMN.west);

\node[vec2] (o11)   at (1.8,0)    {};
\node[vec2] (o12)   at (1.8,-1.1) {};
\node        (o1vd) at (1.8,-2.2) {$\vdots$};
\node[vec2] (o1N)   at (1.8,-3.3) {};

\node        (ovd)  at (1.8,-4.9) {$\vdots$};

\node[vec2] (oMNm1) at (1.8,-6.2) {};
\node[vec2] (oMN)   at (1.8,-7.3) {};

\foreach \n in {o11,o12,o1N,oMNm1,oMN} {
  \draw[thick] ($(\n.north west)+(0,-2mm)$) -- ($(\n.north east)+(0,-2mm)$);
}

\draw[thick, shorten <=1.0mm, shorten >=1.0mm] (f11.east)   -- (o11.west);
\draw[thick, shorten <=1.0mm, shorten >=1.0mm] (f12.east)   -- (o12.west);
\draw[thick, shorten <=1.0mm, shorten >=1.0mm] (f1N.east)   -- (o1N.west);
\draw[thick, shorten <=1.0mm, shorten >=1.0mm] (fMNm1.east) -- (oMNm1.west);
\draw[thick, shorten <=1.0mm, shorten >=1.0mm] (fMN.east)   -- (oMN.west);

\node[rep] (h11)   at (5.8,0)    {$h_{1,1}$};
\node[rep] (h12)   at (5.8,-1.1) {$h_{1,2}$};
\node       (h1vd) at (5.8,-2.2) {$\vdots$};
\node[rep] (h1N)   at (5.8,-3.3) {$h_{1,N}$};

\node       (hvd)  at (5.8,-4.9) {$\vdots$};

\node[rep] (hMNm1) at (5.8,-6.2) {$h_{M,N-1}$};
\node[rep] (hMN)   at (5.8,-7.3) {$h_{M,N}$};

\draw[->, thick, dashed, shorten <=2.5mm, shorten >=2.5mm] (o11.east)   -- (h11.west);
\draw[->, thick, dashed, shorten <=2.5mm, shorten >=2.5mm] (o12.east)   -- (h12.west);
\draw[->, thick, dashed, shorten <=2.5mm, shorten >=2.5mm] (o1N.east)   -- (h1N.west);
\draw[->, thick, dashed, shorten <=2.5mm, shorten >=2.5mm] (oMNm1.east) -- (hMNm1.west);
\draw[->, thick, dashed, shorten <=2.5mm, shorten >=2.5mm] (oMN.east)   -- (hMN.west);

\node[font=\itshape, align=center] at (3.25,-2.25) {remove\\final layer};

\node[fusion] (F) at (8.1,-3.65) {MLP};

\draw[thick, shorten <=1.0mm, shorten >=1.0mm] (h11.east)   -- (F.west |- h11.east);
\draw[thick, shorten <=1.0mm, shorten >=1.0mm] (h12.east)   -- (F.west |- h12.east);
\draw[thick, shorten <=1.0mm, shorten >=1.0mm] (h1N.east)   -- (F.west |- h1N.east);
\draw[thick, shorten <=1.0mm, shorten >=1.0mm] (hMNm1.east) -- (F.west |- hMNm1.east);
\draw[thick, shorten <=1.0mm, shorten >=1.0mm] (hMN.east)   -- (F.west |- hMN.east);

\node[vec5] (pvec) at (9.35,-3.655) {};
\draw[thick] ($(pvec.north west)+(0,-2mm)$)  -- ($(pvec.north east)+(0,-2mm)$);
\draw[thick] ($(pvec.north west)+(0,-4mm)$) -- ($(pvec.north east)+(0,-4mm)$);
\draw[thick] ($(pvec.north west)+(0,-6mm)$) -- ($(pvec.north east)+(0,-6mm)$);
\draw[thick] ($(pvec.north west)+(0,-8mm)$) -- ($(pvec.north east)+(0,-8mm)$);

\draw[thick, shorten <=1.0mm, shorten >=1.0mm] (F.east) -- (pvec.west);

\end{tikzpicture}%
}
\caption{Overview of the expert decomposition and representation-level fusion.}
\label{fig:expert_fusion_overview}
\end{figure}
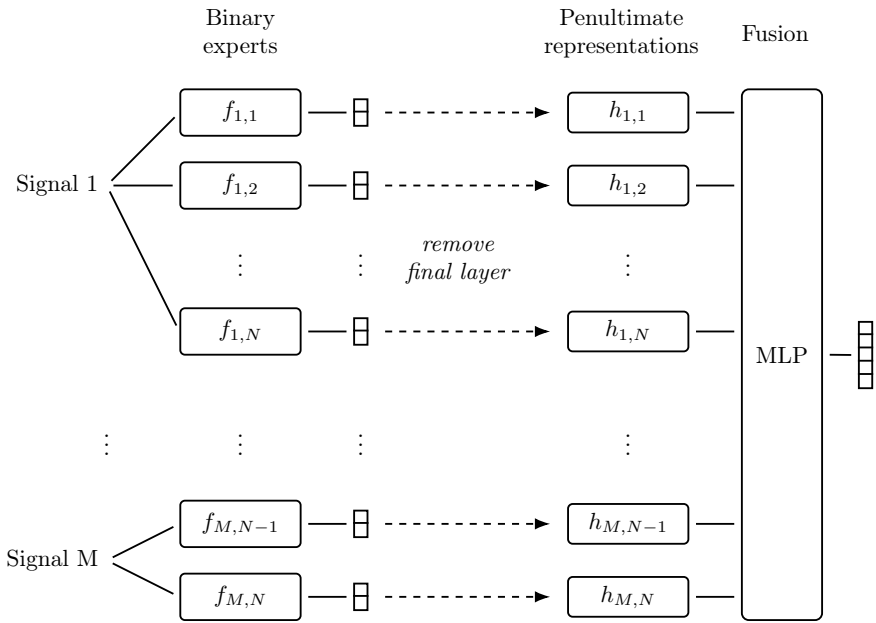

To make modality structure and class-conditional evidence explicit, we reduce the multiclass problem to a set of one-vs-rest binary tasks as shown in Figure \ref{fig:expert_fusion_overview}. For each modality--class pair $(m,c)$, we train a binary expert
\[
f_{m,c} : x_m \mapsto \hat{p}_{m,c} \in [0,1],
\]
where $\hat{p}_{m,c} \approx p(y=c \mid x_m)$ and the target is $y=c$ versus $y\neq c$. Experts (binary classification models such as recurrent neural networks) are trained independently using a binary classification loss, with class imbalance handled as needed (e.g., class-weighted binary cross-entropy or focal loss). This yields $M \times N$ trained experts $\{f_{m,c}\}$.
The baseline expert model (before neural architecture search is performed) is a 1D convolutional residual-Inception network, with a convolutional stem followed by parallel multi-scale temporal branches, normalization, GELU activations, dropout, and global average pooling before a final linear output layer.

Each expert exposes a penultimate representation by removing the final classification layer. We write each expert as
\[
f_{m,c}(x_m) = \sigma\!\left(w_{m,c}^{\top} h_{m,c}(x_m) + b_{m,c}\right),
\]
where $h_{m,c}(x_m) \in \mathbb{R}^{D_{m,c}}$ is the penultimate embedding, $(w_{m,c}, b_{m,c})$ are the parameters of the final linear layer, and $\sigma(\cdot)$ is the logistic sigmoid function. In the fusion stage, we discard the final layer and retain $h_{m,c}(x_m)$ as a learned, modality- and class-conditional representation.

We form a joint representation by concatenating all expert embeddings:
\[
z(x_1,\dots,x_M) = [\,h_{1,1}(x_1);\dots;h_{1,N}(x_1);\dots;h_{M,1}(x_M);\dots;h_{M,N}(x_M)\,].
\]
Then, a fusion model $g(\cdot)$ (a multilayer perceptron) maps $z$ to multiclass logits $\ell \in \mathbb{R}^{N}$, followed by a softmax:
\[
\ell = g(z), \qquad \hat{y} = \arg\max_{c} \mathrm{softmax}(\ell)_c.
\]
Our default training protocol is modular and performed in two stages: experts are trained first and then frozen, after which only the fusion network is trained. Intuitively, each expert $f_{m,c}$ learns what modality $m$ has to say about whether an example belongs to class $c$. Because the expert is trained only on the binary question ``is this class $c$ or not?'' its internal representation is trained to encode class-conditional evidence specific to both the modality and the target class. For example, in sleep-stage classification, an EMG expert for REM may learn patterns associated with reduced muscle tone, whereas an EOG expert for REM may learn to recognize rapid eye movements. The expert embedding $h_{m,c}(x_m)$ can therefore be interpreted as a modality-specific summary of evidence for class $c$, rather than as a generic representation shared across all classes.

\subsection{LLM-guided neural architecture search for expert models}
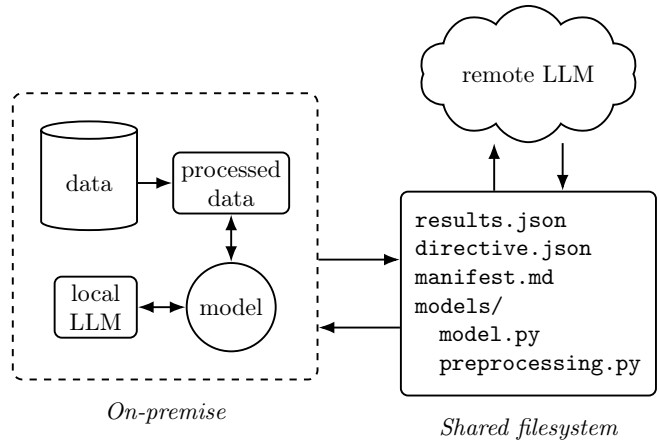
\begin{figure}[htbp]
\centering
\resizebox{0.70\textwidth}{!}{%
\begin{tikzpicture}[
  font=\rmfamily,
  >=Latex,
  node distance=4mm and 5mm,
  box/.style={draw, thick, rounded corners=3pt, minimum width=12mm, minimum height=6mm, align=center},
  circ/.style={draw, thick, circle, minimum size=13mm, align=center},
  filebox/.style={draw, thick, rounded corners=3pt, minimum width=32mm, minimum height=30mm, align=left, inner sep=6pt},
  cloudnode/.style={draw, thick, cloud, cloud puffs=10, cloud puff arc=120, aspect=2,
                    minimum width=30mm, minimum height=13mm, align=center},
  note/.style={font=\bfseries}
]

\node[draw, thick, cylinder, shape border rotate=90, aspect=0.28,
      minimum height=16mm, minimum width=14mm, align=center] (data) {data};

\node[box, right=of data] (proc) {processed\\data};

\node[circ, below=7mm of proc] (model) {model};

\node[box, left=7mm of model] (llm) {local\\LLM};

\draw[->, thick] (data.east) -- (proc.west);

\draw[<->, thick] (proc.south) -- (model.north);

\draw[<->, thick] (llm.east) -- (model.west);

\node[draw, thick, dashed, rounded corners=4pt,
      fit=(data)(proc)(llm)(model),
      inner sep=4mm] (onprem) {};

\node[font=\itshape, below=2mm of onprem.south, anchor=north] {On-premise};

\node[filebox, anchor=west] (fs) at ($(model.east)+(18mm,2mm)$)
{
\texttt{results.json}\\
\texttt{directive.json}\\
\texttt{manifest.md}\\
\texttt{models/}\\
\hspace{1em}\texttt{model.py}\\
\hspace{1em}\texttt{preprocessing.py}
};

\node[font=\itshape, below=2mm of fs] {Shared filesystem};

\node[cloudnode, above=7mm of fs] (agent) {remote LLM};

\draw[->, thick] ([xshift=-5mm]fs.north) -- ([xshift=-5mm]agent.south);
\draw[->, thick] ([xshift= 5mm]agent.south) -- ([xshift= 5mm]fs.north);

\draw[->, thick] ($(onprem.east |- fs.west) + (0,5mm)$) -- ($(fs.west) + (0,5mm)$);
\draw[->, thick] ($(fs.west) + (0,-5mm)$) -- ($(onprem.east |- fs.west) + (0,-5mm)$);

\end{tikzpicture}%
}
\caption{Schematic of the data-local autonomous NAS loop. The data and all training remain on-premise. The local execution (model debugging, training and hyperparameter optimization) consumes directives and writes results to a shared filesystem. A remote LLM interacts only through these non-sensitive files. A smaller, local LLM debugs simple errors in the code proposed by the remote LLM, since error messages may contain sensitive data.}
\label{fig:system_overview}
\end{figure}

Neural architecture search is applied to the individual expert models $\{f_{m,c}\}$. We implement architecture search as an agentic loop in which a remote LLM serves as the \emph{search controller}, proposing candidate expert architectures based on observed fitness trends (validation performance and learning dynamics) accumulated across prior evaluations. All training and evaluation are executed locally under a fixed protocol and recorded as structured artifacts. Figure \ref{fig:system_overview} shows the NAS loop.

The search space is intentionally flexible. Rather than selecting from a small hand-designed menu of templates, the remote controller is allowed to propose substantial changes to the expert pipeline, including preprocessing, feature-extractor structure, layer widths and depths, kernel configurations, normalization, pooling, dropout and other hyperparameters. In practice, the remote LLM proposes architecture and preprocessing code, while the local executor runs all candidates under a fixed evaluation protocol. Thus, the controller has broad freedom at the level of model design, but not at the level of data access.

The controller receives only non-sensitive summaries from the local environment, such as architecture descriptors, hyperparameters, training budgets, validation metrics, optionally learning-curve summaries, runtime, and failure traces. The controller does not access raw data, examples, or any dataset-derived artifacts beyond aggregated metrics. The controller then outputs a new architecture, which is then evaluated locally.

Communication between the remote controller and the on-premise executor is mediated by files in a shared filesystem. 
Human-readable files such as \texttt{manifest.md} preserve the current search context, recent findings, and experiment rationale across controller calls, whereas structured JSON files such as \texttt{results.json} define executable instructions and machine-readable outputs. 
In particular, the controller writes a structured directive specifying which binary experts to target and which candidate preprocessing and model architecture to use. 
The local executor validates this directive, loads the corresponding candidate code, performs preprocessing, trains and evaluates the requested experts, and writes back structured results containing validation metrics, learning-curve summaries, runtime, and failure traces. 
This artifact-based interface makes the search loop restartable and auditable: every proposal, execution, outcome, and repair attempt is preserved in the experimental record. It also ensures that the remote LLM never has access to the dataset.

Given a proposed architecture for an expert $f_{m,c}$, the local executor trains the corresponding binary classifier under a fixed training protocol and computes fitness on a held-out validation split using an appropriate binary metric (e.g., AUC, F$_1$, or balanced accuracy). The evaluation protocol (data splits, loss definition, optimization settings, and training budget) is fixed for the duration of a search run to ensure comparability across proposals. The resulting fitness and auxiliary summaries are appended to the search history and returned to the controller.

The overall procedure is orchestrated as a persistent research cycle that repeatedly (1) proposes expert architecture candidates, (2) instantiates and trains candidates locally, (3) evaluates and ranks results, and (4) records all relevant artifacts. This design supports post-hoc auditing since the full sequence of proposals and outcomes can be reconstructed from the artifact trail.

In deployments with data-locality constraints, dataset access and model training remain strictly on-premise. The remote controller interacts with the local environment only through the constrained artifact interface described above, ensuring that sensitive data are not exposed through prompts or logs while still enabling controller-driven adaptation based on fitness trends.

\subsection{Experimental setup and benchmarks}

We evaluate our method on two regimes. Firstly, we use the UEA30 dataset, a widely used benchmark for multivariate, multiclass time-series classification \cite{ruiz_great_2021}. 
Details about the datasets are reported in Table \ref{tab:dataset_results_summary}.
Secondly, we use the SleepEDFx dataset, a sleep staging benchmark dataset with heterogeneous clinical modalities such as EEG, EOG and EMG.  This data set is available through Physionet and consists of polysomnography recordings of healthy adults \cite{kemp2000analysis}. 

We evaluated three configurations: 
\begin{enumerate}
    \item
 \textbf{Baseline end-to-end:} A baseline trained directly on the multiclass task in an end-to-end approach, 
 \item
 \textbf{Baseline staged:} A baseline in which binary experts were trained individually and then fused as described in section \ref{sec:expert}, and 
\item
 \textbf{LLM NAS:} the proposed staged system with LLM-guided NAS for a binary expert model architecture.
\end{enumerate}
For each dataset, train/validation/test splits were fixed for the duration of an experiment. For comparison to prior work, we report a reference range of the accuracy of methods from the literature. For the UEA30 datasets, this range was constructed as the minimum and maximum test accuracy reported across three recent benchmark studies spanning 2021–2025 \cite{ruiz_great_2021,ismail-fawaz_look_2025,foumani_improving_2024}. 
We selected studies that covered the UEA archive broadly, rather than papers reporting results on only one or a few datasets, so that the comparison would reflect methods evaluated across the benchmark suite rather than dataset-specific tuning.
For SleepEDFx, the reference range was constructed from recent studies reporting deep learning 5-class sleep staging on SleepEDFx \cite{zhu_convolution-_2020,joe_classification_2022,huang_unified_2026,guo_flexsleeptransformer_2024}.
The implementation of our methods can be found at \href{https://github.com/emilhar/arl}{https://github.com/emilhar/arl}.

\section{Results}\label{sec:results}

\subsection{Experimental overview}

Table \ref{tab:dataset_results_summary} summarizes dataset characteristics and test set accuracy for three configurations: (1) the baseline model trained end-to-end, (2) the baseline model trained in stages, with the binary experts trained individually and then fused, and (3) the LLM-guided NAS model.
Across the benchmarks and tested configurations, we make two broad findings:
First, the staged expert model is already a strong baseline and often improves upon direct end-to-end training.
Second, the LLM-guided NAS can further improve the baseline, although gains depend on the dataset.
Unless otherwise stated, the NAS was run with a 10-cycle budget.
Longer searches were performed on two illustrative cases: SleepEDFx, where the NAS method is evaluated on a multimodal clinical problem, and EthanolConcentration, where the baseline model leaves substantial room for improvement.

\begin{table}[h!]
\centering
\caption{Benchmark overview and performance summary. For each dataset, the table reports the number of modalities ($m$), number of variates ($d_m$), sequence length ($T_m$), number of classes ($C$), train and test set sizes, and test accuracy (\%) for three configurations: the end-to-end baseline, the staged baseline with individually trained binary experts followed by fusion, and the proposed LLM-guided NAS model. The final column reports the range of accuracies reported in recent published work for the respective dataset.
Unless otherwise stated, NAS was run with a budget of 10 cycles.
``--'' indicates that the configuration was not evaluated, either because the dataset was too small or because the baseline model achieved 100\% accuracy on the validation set. 10 cycle searches unless stated otherwise.}
\label{tab:dataset_results_summary}
\scriptsize
\setlength{\tabcolsep}{3.5pt}
\renewcommand{\arraystretch}{1.15}
\resizebox{\textwidth}{!}{%
\begin{tabular}{lcccccccccc}
\toprule
\rule{0pt}{0.7cm}
Dataset
& \rot{Mod, $m$}
& \rot{Dim, $d_m$}
& \rot{Len, $T_m$}
& \rot{Class, $C$}
& \rot{Train size}
& \rot{Test size}
& \makecell[b]{Baseline\\end-to-end}
& \makecell[b]{Baseline\\staged}
& \makecell[b]{LLM \\ NAS }
& \makecell[b]{Previous\\work} \\
\midrule

SEDF & 3 & 5 & 3000 & 5 & 112,287 & 10,774 & 82.9 & 84.7 & 87.9$^{a}$ & [82.8 - 94.0] \\ 
AWR  & 1 & 9 & 144 & 25 & 275 & 300 & 91.7& 96.3& --$^{b}$& [94.3 - 99.6]\\ 
AF   & 1 & 2 & 640 & 3 & 15 & 15 & 33.3& 33.3& --& [6.7 - 74.0]\\ 
BM   & 2 & 6 & 100 & 4 & 40 & 40 & 100& 92.5& --$^{b}$& [72.2 - 100.0]\\ 
CT   & 1 & 3 & 0 & 20 & 1422 & 1436 & 90.1& 95.8& 95.8& [98.7 - 99.6]\\ 
CR   & 1 & 6 & 1197 & 12 & 108 & 72 & 95.8& 93.1& --$^{b}$& [93.1 - 100.0]\\ 
ER  & 1 & 4 & 65 & 6 & 30 & 270 & 90.7& 85.2& 80.1& [82.4 - 98.9]\\ 
EW   & 1 & 6 & 17984 & 5 & 128 & 131 & 75.6& 77.8& 84.0& [42.0 - 95.4]\\ 
EP   & 1 & 3 & 206 & 4 & 137 & 138 & 96.4& 97.1& --$^{b}$& [67.0 - 100.0]\\ 
EC   & 1 & 3 & 1751 & 4 & 261 & 263 & 28.1& 32.7& 44.1$^{c}$& [27.8 - 82.4]\\ 
HMD   & 1 & 10 & 400 & 4 & 160 & 74 & 40.5 & 20.3 & 21.6 & [21.6 - 59.5]\\ 
HW   & 1 & 3 & 152 & 26 & 150 & 850 & 13.1& 14.0 & 14.0 & [18.0 - 65.7]\\ 
JV  & 1 & 12 & 29 & 9 & 270 & 370 & 96.5 & 97.0 & 94.9 & [91.4 - 98.9] \\ 
LIB  & 1 & 2 & 45 & 15 & 180 & 180 & 60.6 & 76.7 & 78.9 & [75.6 - 94.1] \\ 
LSST & 1 & 6 & 36 & 14 & 2459 & 2466 & 49.1 & 66.5 & 66.5 & [34.0 - 66.4] \\ 
NAT  & 1 & 24 & 51 & 6 & 180 & 180 & 88.9 & 84.7 & 90.1 & [76.1 - 97.1] \\ 
PD   & 1 & 2 & 8 & 10 & 7494 & 3498 & 96.2 & 95.7 & 95.7 & [87.5 - 99.7] \\ 
PS   & 1 & 11 & 217 & 39 & 3315 & 3353 & 23.6& 26.2& 26.2& [10.2 - 36.7] \\ 
RS   & 2 & 6 & 30 & 4 & 151 & 152 & 79.6& 75.7& 73.0 & [73.7 - 92.8] \\ 
SAD  & 1 & 13 & 93 & 10 & 6599 & 2199 & 95.2& 98.5& 98.5 & [98.3 - 99.5] \\ 
SWJ  & 1 & 4 & 2500 & 3 & 12 & 15 & 33.3& 33.3& -- & [22.0 - 66.7] \\ 
UWG  & 1 & 3 & 315 & 8 & 120 & 320 & 49.7& 83.1& 83.1 & [71.1 - 94.4] \\ 
\bottomrule
\end{tabular}%
}
\vspace{1mm}
\parbox{\textwidth}{\scriptsize 
$^{a}$ 30 cycle search.\\
$^{b}$ NAS was automatically skipped because validation set showed 100\% accuracy.\\
$^{c}$ 60 cycle search.\\
Abbreviations: SEDF (SleepEDFx), AWR (ArticularyWordRecognition), AF (AtrialFibrillation), BM (BasicMotions), CT (CharacterTrajectories), CR (Cricket), ER (ERing), EW (EigenWorms), EP (Epilepsy), EC (EthanolConcentration), HMD (HandMovementDirection), HW (Handwriting), JV (JapaneseVowels), LIB (Libras), LSST (LSST), NAT (NATOPS), PD (PenDigits), PS (PhonemeSpectra), RS (RacketSports), SAD (SpokenArabicDigits), SWJ (StandWalkJump), and UWG (UWaveGestureLibrary).
}

\end{table}

\subsection{Baseline model}

The staged baseline performs competitively before LLM-guided NAS is applied as can be seen in Table \ref{tab:dataset_results_summary}.
On SleepEDFx, the staged baseline model reaches 84.7\% accuracy, falling within the range reported in prior work (82.8-94.0). 
As seen across the UEA datasets, training the baseline model in stages improves on the end-to-end baseline in 12 of 21 cases, suggesting that training the binary experts individually before fusion is often, but not uniformly, beneficial.

\subsection{Effect of LLM-guided NAS}

The effect of the LLM-guided NAS is mixed across datasets.
The clearest gain, relative to the range of previous work, is observed in SleepEDFx, where a 30-cycle NAS increases the test accuracy from 84.7\% to 87.9\%.
This moves the model well into the state-of-the-art range found in recently published results and beyond inter-scorer variability among human sleep scorers \cite{fiorillo_automated_2019}.
This shows that the LLM-guided NAS can make meaningful improvements to an already competitive baseline model.
A second case where the benefits of the LLM-guided NAS are apparent is EthanolConcentration (EC) dataset, where a longer, 60-cycle search improves accuracy from 32\% to 44.1\%.
In contrast to SleepEDFx, in this dataset, the search starts from a relatively weak baseline, suggesting that the search procedure can recover performance even when the initial expert architecture is poorly matched to the task.
Looking at Table \ref{tab:dataset_results_summary}, the effect of the LLM-guided NAS is dependent on the specific dataset.

More modest improvements are observed on several other datasets, including EigenWorms (EW, 77.8\% to 84.0\%), Libras (LIB, 76.7\% to 78.9\%), and NAT (NATOPS, 84.7\% to 90.1\%).
However, several datasets show no improvement, while a few degrade slightly, likely due to overfitting.
This suggests that the LLM-guided NAS is most useful when the baseline model architecture is sub-optimal for the dataset.

\begin{figure}[htbp]
\centering
\includegraphics[width=0.79\textwidth]{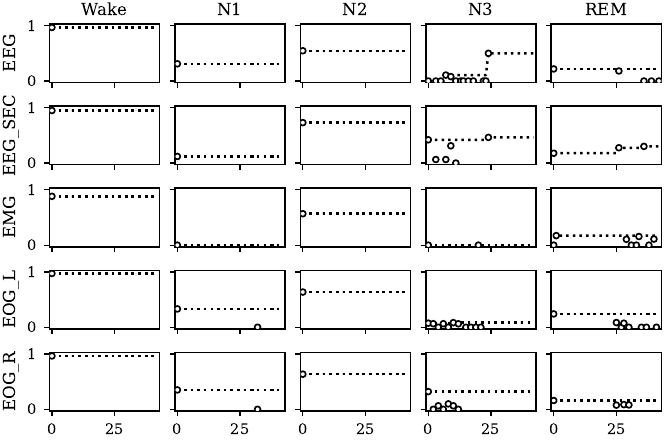}
\caption{Search trajectories for binary experts during a 30-cycle NAS on the SleepEDFx dataset. Each panel corresponds to one class-modality expert. Dashed lines show the F$_1$ score of the best found expert, and circles show the F$_1$ score of the expert trained during the cycle.}
\label{fig:nas_sleepedfx}
\end{figure}

\begin{figure}[htbp]
\centering
\includegraphics[width=0.75\textwidth]{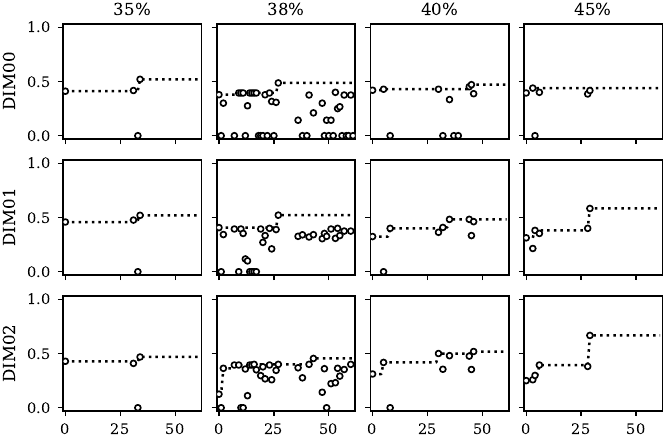}
\caption{Search trajectories for binary experts during a 60-cycle NAS on the EthanolConcentration dataset. Each panel corresponds to one class-modality expert. Dashed lines show the F$_1$ score of the best found expert, and circles show the F$_1$ score of the expert trained during the cycle.}
\label{fig:nas_ethanol}
\end{figure}

Figures 3 and 4 show the search dynamics of the LLM-guided NAS.
Each panel corresponds to one class-dimension expert. 
Circles show the F1 score obtained when that expert was trained in a given cycle, while dashed lines show the best F$_1$ achieved so far for that expert.
Because only one class-modality architecture is proposed per cycle, many panels show no change for extended periods, and flat lines indicate either that the expert was not examined or that retraining did not improve on the current best result.
Figure \ref{fig:nas_sleepedfx} shows that on SleepEDFx, gains are concentrated on the weakest experts (N3 and REM sleep), whereas in Figure \ref{fig:nas_ethanol} we see improvements across all experts.

\subsection{Overall takeaways}

These results support three main conclusions.
First, the staged expert-fusion baseline model is already a strong model, fit for a broad set of multivariate multiclass time-series classification tasks.
Second, the LLM-guided NAS can provide additional gains, with the largest improvements in cases where the baseline architecture is a poor fit or where complex, multimodal data leaves room for targeted expert adaptation.
Third, the practical value of the proposed framework extends beyond benchmark accuracy alone. Instead, it lies in enabling automated, long-horizon architecture search in sensitive, practical situations, where raw data cannot be exposed to an external LLM.

\section{Discussion}\label{sec:discussion}

\subsection{Open Access Tool}

The proposed framework is accessible as an open-source software repository under the MIT license.
Our development and open-access sharing of this tool enable ML researchers to systematically and automatically design models for multimodal time-series classification. Furthermore, the tool's design is sufficiently interpretable, allowing domain experts with no in-depth ML knowledge to apply these methods. A core strength of our method is that the user can understand what modality and class is being worked on and see the reasoning behind the LLM's model suggestion. In the case of sleep staging, for example, during one cycle of our tests the system reported that REM was the weakest ensemble class and that improving it would lift macro-F1 the most. The LLM controller further pointed out that EMG expert had zero F1 for REM, even though REM is characterized by low EMG tone and atonia, and consequently proposed that simple amplitude features could unlock strong precision gains. Since this information is shared in a human-readable way, a domain expert can monitor the process and use their domain knowledge to control the model development, without requiring in-depth ML knowledge to implement the changes. 

On the one hand, this more intuitive human approach allows people with in-depth knowledge of the data to apply ML, which may lead to new perspectives and contributions. Our results are in line with Miguel-Morante et al., who argue that this human interpretability adds value to the system \cite{miguel2025integrating}. On the other hand, there is a risk associated with making ML accessible to people without in-depth ML knowledge \cite{zoller2021benchmark}. This could lead to applications that are not properly validated, overfitting, or being misinterpreted, potentially resulting in unreliable or clinically misleading outcomes. As Somer points out in their review from 2025, there are ways to mitigate the risk of domain experts using simplified tools to create predictive models, such as setting up an accountability framework, offering special training to domain experts, or increasing the explainability of the model \cite{somer2025algorithmic}.

\subsection{Data Privacy}

This paper showed that developing a model that stores data and trains locally can still achieve performance comparable to or better than that of common benchmarking datasets for multimodal time series classification. Our system architecture, therefore, adheres to the concept of privacy by design (PBD), which centers on integrating privacy-preserving design into a system's architecture rather than adding privacy measures afterward \cite{herath2024data}. Our architecture furthermore strongly aligns with the PBD design principles defined by Cavoukian et al. \cite{cavoukian2009privacy} by i.) enacting privacy proactively, ii.) preserving data privacy by default, and iii.) preserving full functionality of the system.

\subsection{Limitations}

A practical evaluation challenge arises when LLMs are used to assist with planning, debugging, and documenting experiments. Canonical NAS benchmarks and leaderboards are heavily documented, and an assistant model may reproduce benchmark-specific architectures and recipes that are widely available in the public literature and online. This aligns with the work of Tornede et al., who note that the inability to trace which data the LLM used for training is one of the most severe risks in using and evaluating LLMs for AutoML \cite{tornede2023automl}.
This makes it difficult to disentangle performance gains attributable to the proposed search-and-evaluate mechanism from gains due to benchmark-specific prior art. For this reason, we emphasize \emph{domain-representative} multiclass multimodal time-series tasks under data-local constraints, where the objective is robust, auditable experimental progress rather than leader board ranking.

\section{Conclusion}\label{sec:conclusion}
In this paper, we presented a novel framework for a data-local autonomous NAS guided by an LLM to tackle the challenging problem of multiclass multimodal time-series classification. Firstly, we turned the classification problem into a fusion network of binary expert classifiers, each of whom trains on a single modality. This allows the model to focus on the properties of each data stream in isolation. The output of the experts is then fused together to form the final classifier. As our results on 22 benchmark datasets demonstrate, this modification already brings significant performance gains. Secondly, we implemented an LLM-guided NAS that had the objective of improving the experts by proposing changes to their model architecture. 
The NAS trained over multiple cycles and was able to improve the architecture for several of the benchmark datasets, leading to an even greater enhancement in performance in some cases.
This LLM-guided NAS, while utilizing the power of modern LLMs, is data privacy preserving, as the models are trained locally, making this a feasible approach for developing ML models in domains such as healthcare.  

Although our experiments are limited, our work shows great potential and opens several doors for future work. On the one hand, the fusion network of binary experts could be adapted to regression problems and forecasting, which are common when working with time series, making this modification highly relevant. In this work, we focus on multivariate time series, but other data modalities, such as images and text, could be included as well to make the framework more generalized. On the other hand, the LLM-guided NAS itself is a novel approach to finding optimal architectures in neural networks, and our prototype is a simple one. With more sophisticated LLMs and prompting, we will focus our future research on faster and more effective learning.

\section*{Acknowledgments}
 This research was funded by the Icelandic Research Fund under Grant 2410607-051. 
LLMs were used during the writing of this paper to enhance the readability of the text.

The authors have no competing interests to declare that are relevant to the content of this article.

\bibliographystyle{splncs04}
\bibliography{refs}

\end{document}